\documentclass{article}

\usepackage{PRIMEarxiv}

\usepackage[utf8]{inputenc} % allow utf-8 input
\usepackage[T1]{fontenc}    % use 8-bit T1 fonts
\usepackage{hyperref}       % hyperlinks
\usepackage{url}            % simple URL typesetting
\usepackage{booktabs}       % professional-quality tables
\usepackage{amsfonts}       % blackboard math symbols
\usepackage{nicefrac}       % compact symbols for 1/2, etc.
\usepackage{microtype}      % microtypography
\usepackage{lipsum}
\usepackage{fancyhdr}       % header
\usepackage{graphicx}       % graphics
\usepackage{amsmath}
\usepackage{placeins}

\usepackage{geometry} 
\usepackage{float}
\usepackage{tabularx} 

\graphicspath{{media/}}     % organize your images and other figures under media/ folder 

\title{conserving human creativity with evolutionary generative algorithms: a case study in music generation
}

\author{
  Justin Kilb \\
  Ridge AI \\
  Wondr Search \\
   \texttt{\ wondrsearch@gmail.com} \\
   \And
  Caroline Ellis \\
  Ridge AI \\
  Wondr Search \\
  \texttt{\ wondrsearch@gmail.com} \\
}

\begin{document}
\maketitle

\begin{abstract}

This study explores the application of evolutionary generative algorithms in music production to preserve and enhance human creativity. By integrating human feedback into Differential Evolution algorithms, we produced six songs that were submitted to international record labels, all of which received contract offers. In addition to testing the commercial viability of these methods, this paper examines the long-term implications of content generation using traditional machine learning methods compared with evolutionary algorithms. Specifically, as current generative techniques continue to scale, the potential for computer-generated content to outpace human creation becomes likely. This trend poses a risk of exhausting the pool of human-created training data, potentially forcing generative machine learning models to increasingly depend on their random input functions for generating novel content. In contrast to a future of content generation guided by aimless random functions, our approach allows for individualized creative exploration, ensuring that computer-assisted content generation methods are human-centric and culturally relevant through time.

\end{abstract}

\section{Introduction}

Music generation, a field that intersects art and Artificial Intelligence (AI), has employed various methodologies, including stochastic modeling, machine learning (ML), and evolutionary optimization techniques. Hidden Markov Models have been utilized for their probabilistic nature in generating sequences, including musical notes \cite{davismoon2010combining} \cite{mcvicar2014autoleadguitar}. Artificial neural networks have gained prominence due to their ability to learn temporal dependencies, making them suitable for sequence generation tasks such as music composition  \cite{eck2002first} \cite{makris2017combining} \cite{abbou2020deepclassic}  \cite{mishra2019long}. 

Additionally, Evolutionary Algorithms (EAs) have emerged as a powerful tool for the generation of music \cite{horner1991genetic} \cite{jacob1995composing} \cite{Anderling2014}. Unlike ML techniques, which often rely on large datasets to learn and replicate existing musical content, EAs can operate by learning from a single user's feedback \cite{tokui2000music}. 

Hybrid approaches generate music by combining genetic algorithms for note generation with neural networks for evaluating musical fitness, suggesting improved music patterns through, for example, tournament selection and two-point crossover techniques \cite{doush2020automatic}. Similarly, Bi-LSTM neural networks have been utilized in the scoring and modeling process \cite{farzaneh2019music}.

However, hybrid methodologies that utilize EAs and ML rely on pre-existing musical data for training models, which has created friction for musicians and listeners, as it blurs the lines between originality and replication. Consequently, both artists and audiences are likely to question the authenticity and creativity in a future where the boundaries of human and machine-generated art are increasingly indistinct. 

This tension highlights the potential for generative content methods that do not depend on pre-existing works, but instead evolve creatively from the preferences of a single user. Unlike ML, EAs retain the fundamental characteristics of human creativity, its inherent variability, while also offering robust tools for artistic enhancement. The EA serves as a facilitator, enabling the exploration and refinement of creativity while capturing the unique artistic expression of a single individual.

\subsection{Creativity fueled by mutation}

Human creativity is deeply rooted in personal experiences, cultural backgrounds, and environmental contexts \cite{shao2019how}. These factors introduce diversity into the creative process, resulting in a wide array of artistic expressions. When employing math and computer science to augment human creativity, it is essential to preserve these natural variations. Optimization techniques, particularly those involving EAs, are designed to replicate these variations through mechanisms such as mutation and crossover. These methods introduce stochasticity into the optimization process, akin to the spontaneous and unpredictable nature of human creativity that evolves over time.

As O’Toole and Horvát demonstrated, the introduction of novelty in music is crucial for both commercial success and cultural evolution. Their analysis revealed that songs with an optimal level of differentiation from their contemporaries are more likely to achieve commercial success. The most popular songs exhibit a balance of novelty, standing out enough to be distinct but not so much that they become unrelatable to listeners \cite{otoole2023novelty}. This concept of "optimal differentiation" suggests that successful music must innovate while maintaining a connection to established norms. Furthermore, their findings highlight that the relationship between novelty and success is not linear; overly novel songs tend to perform poorly, emphasizing the importance of achieving the right degree of novelty.

This insight underscores the potential of incorporating mutation and crossover functions in EAs to continually introduce new variations in music generation. By maintaining a dynamic balance between human-guided mutation and the explorative scale of computers, these algorithms can potentially emulate the natural mutations and environmental influences that historically drive the evolution of music genres. This approach increases the likelihood that computer-assisted content generation remains innovative, culturally relevant, and human centric.

\subsection{Human-guided mutation}

Current generative AI methods aim to introduce randomness to explore new types of content, but they are inherently limited by the characteristics of their training data. This limitation is, generally, intentional, as the goal is to replicate the training data closely enough for users to recognize the output as the intended category, while still providing enough variation to avoid exact duplicates. This is achieved, for instance, by adding randomness to the noise vectors that modify input data in neural networks. However, this randomness is typically constrained by the distribution of the random function itself. As a result, the diversity of generated content remains confined within the limits of this random function, restricting the potential and efficiency of discovering new generative styles, which may be computationally intensive or lie outside the boundaries of the random distribution. This is also true for other "diversity-adding" techniques like stochastic sampling, latent space exploration, conditional generation, data augmentation, beam search, and regularization. Figure 1 illustrates these constraints, showing how the distribution of the random function applied to the model inputs limits the maximum deviation of outputs.

\begin{figure}[htbp]
  \centering
  \includegraphics[width=0.95\textwidth]{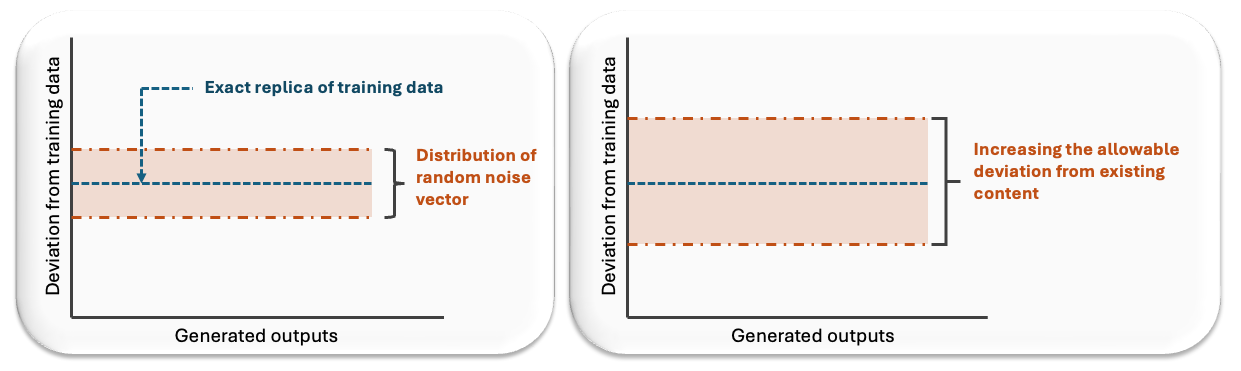}
  \caption{Left: A noise vector creates a distribution of generated content that deviates away from training dataset replication. Right: Increasing the maximum allowable deviation by increasing the bounds of the random distribution in the noise vector.}
  \label{fig:figure1}
\end{figure}

\FloatBarrier % Add this command after the first figure

In contrast to the directionless boundaries of random functions, human creativity, through time, is perhaps, influenced by a wide range of cultural and environmental factors that can significantly alter an individual's biases in ways that likely exceed the static and random bounds of noise vectors. Therefore, while currently-deployed methods can generate outputs with some diversity, they are inherently limited by the scope of their training data and currently lack the adaptive, context-sensitive evolution driven by human experience and societal changes. 

Alternatively, employing EAs with human feedback supports the natural artistic process, characterized by continuous variation and adaptation. This process can be executed through the creation of a computer-generated melody and a human feedback loop. Initially, a melody is generated with constraints informed by music theory but not by pre-existing works. Then, a human assigns a score to the generated melody, and the process iterates. As the human scores the melodies, EAs adapt to retain and enhance the properties that are likely to lead to better scores. 

Consider Figure 2, where we introduce an additional dimension of content variation. The image on the left demonstrates the confined bounds of the random function in a noise vector, which leads to randomly generated content, marked with the blue spheres. In this process, the blue spheres will appear randomly, only within the boundaries of the random function, thus limiting their content variation reach. In the right image however, the blue spheres, which also represent generated content, are directionally guided by the human's score. 

\begin{figure}[htbp]
  \centering
  \includegraphics[width=0.9\textwidth]{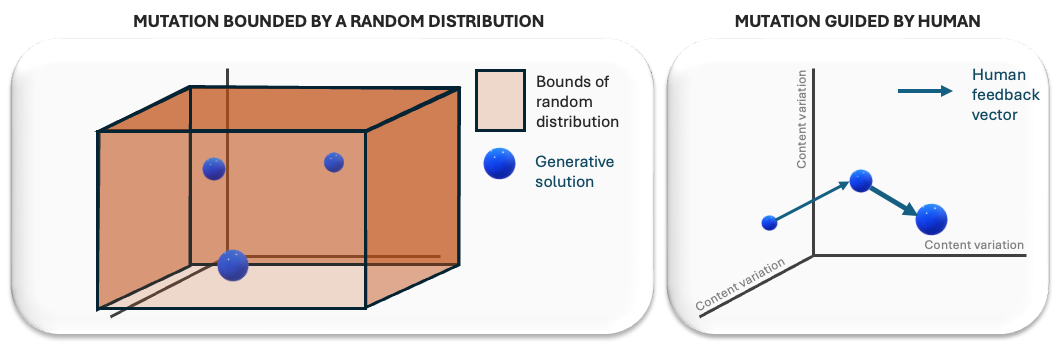}
  \caption{Left: Mutation of new content is bounded by the limits of the random function in a noise vector. Right: Mutation is guided by a human's score.}
  \label{fig:figure1}
\end{figure}

\FloatBarrier % Add this command after the first figure

 The directional guidance of mutation, by a human, could alleviate the otherwise computationally-expensive search procedures to find the desired location of content generation within the random distribution, since the volumetric explorative space of a random functions grows exponentially with an increase in dimensions. Because an EA, human-in-the-loop process is directional, it means that it is possible to efficiently search outside the bounds of a random distribution. This search process can efficiently lead to a great degree of diversity between the preferences of two individuals. Figure 3 illustrates this concept, where the red spheres represent the process of two different individuals guiding mutation to different locations, outside the bounds of a random distribution.  

\begin{figure}[htbp]
  \centering
  \includegraphics[width=0.6\textwidth]{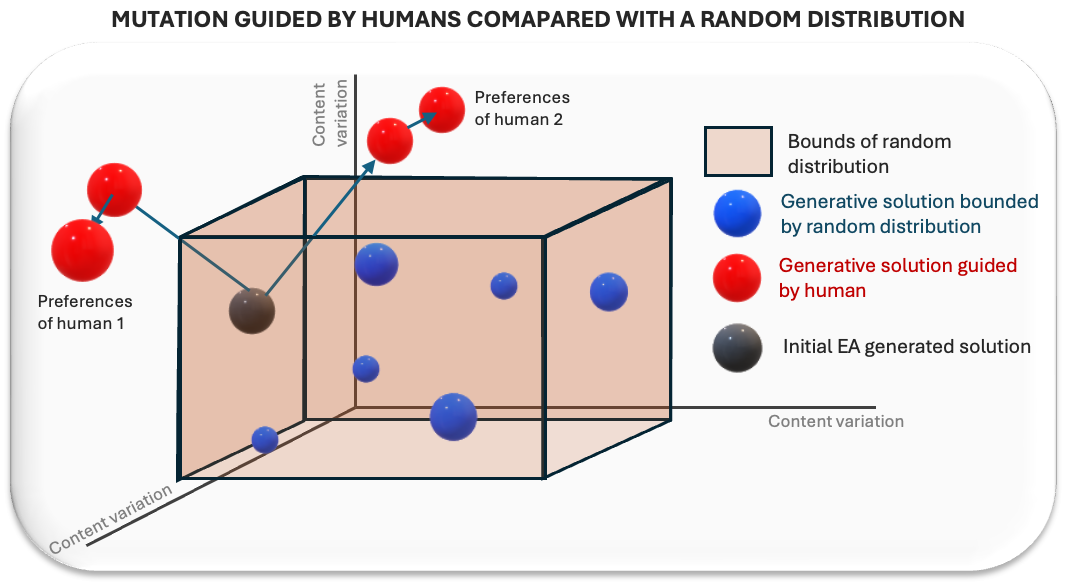}
  \caption{Efficient mutation, guided by human feedback, explores outside the bounds of a random distribution.}
  \label{fig:figure3}
\end{figure}

\FloatBarrier % Add this command after the first figure

Comparing content generation through traditional methods and EAs reveals a fundamental distinction. Traditional ML methods excel in producing close replicas of existing categories, such as generating images of dogs for non-artistic purposes. In these instances, novelty is not typically the goal; instead, the aim is to create recognizable outputs that differ just enough to avoid plagiarism, while reducing cost. This is akin to a business owner who needs a functional website and prioritizes meeting a standard of sufficient quality rather than artistic innovation. For tasks like these, traditional ML methods seem to be an excellent option. However, for creative endeavors where maximizing novelty is desired, traditional ML methods risk creating an asymmetry between human and machine-generated content. Given the enormous and continuously growing scalability of contemporary generative techniques, it is hard to imagine a future where human output surpasses that of computers. Thus, as traditional ML generative models reach their full potential, we risk depleting the reservoir of human-created training data, forcing these models to increasingly rely on their random input functions to achieve novelty. 

\section{Methods}

\subsection{Representation of melody with numbers}

Music can be quantified with mathematical representation. Such mathematical frameworks enable a structured exploration of musical properties and relationships, providing a robust basis for algorithmic composition and analysis \cite{Smith2018}.

The algorithm in this work generates musical pitch, which is represented as sequences of frequencies corresponding to MIDI numbers and musical notes. Duration of notes are represented by fractions of whole notes. Velocity is represented with abrupt signal deviation, rhythm is generated through temporal metrics, and the tempo of the melody is constrained by the beats per minute. More complex features, like aggregate note selection are constrained with frequency bounds in certain key signatures. Finally, note articulation to mimic the dynamics of live instruments is represented with vector time-series modulation or in the post-algorithm live recording of generated melodies with Digital Audio Workstation MIDI instrument modulation. 

By treating melodies as vectors of real numbers, we apply mathematical operations such as generation, mutation, and crossover to explore and refine musical compositions systematically. 

\subsection{Differential Evolution Algorithm}

Although many works in EAs for music generation rely on Genetic Algorithms, we explore the use of Differential Evolution (DE). Initially proposed by Storn and Price in the 1990s, DE is known for its simplicity and efficiency in handling real-valued optimization problems \cite{StornPrice1997}. Unlike genetic algorithms, which rely on binary or discrete representations and traditional crossover and mutation operators, DE operates on real-valued vectors and uses a differential mutation operator that is designed for continuous optimization.

\subsubsection{Initialization}
The DE algorithm begins with a randomly generated initial population of candidate solutions (melodies). Each individual melody in the population is represented as a vector of real numbers:
\[
\mathbf{x}_i^G = [x_{i,1}^G, x_{i,2}^G, \ldots, x_{i,D}^G],
\]
where \(G\) denotes the generation number, \(i\) is the index of the individual, and \(D\) is the dimensionality of the problem.

\subsubsection{Mutation}
Mutation in DE is distinct from traditional genetic algorithms. Instead of random bit flips, DE generates a mutant vector by adding a weighted difference between two randomly selected population vectors to a third vector:
\[
\mathbf{v}_i^{G+1} = \mathbf{x}_{r1}^G + F \cdot (\mathbf{x}_{r2}^G - \mathbf{x}_{r3}^G),
\]
where \(\mathbf{v}_i^{G+1}\) is the mutant vector, \(F\) is the scaling factor, and \(r1, r2, r3\) are distinct indices randomly chosen from the population. This process ensures that the mutation is proportional to the existing differences in the population, promoting smooth transitions and gradual improvements from one generation of melodies to the next.

\subsubsection{Crossover}
The crossover operation creates a trial vector by combining elements from the target vector and the mutant vector. The crossover probability \(C_r\) controls the rate at which parameters are exchanged:
\[
u_{i,j}^{G+1} = \begin{cases} 
v_{i,j}^{G+1} & \text{if } \text{rand}_j \leq C_r \\
x_{i,j}^{G} & \text{otherwise}
\end{cases},
\]
where \(u_{i,j}^{G+1}\) is the trial vector, and \(\text{rand}_j\) is a random number between 0 and 1.

\subsubsection{Selection}
Selection in DE is deterministic. The trial vector is compared to the target vector, and the one with the better fitness value survives to the next generation:
\[
\mathbf{x}_i^{G+1} = \begin{cases} 
\mathbf{u}_i^{G+1} & \text{if } f(\mathbf{u}_i^{G+1}) \leq f(\mathbf{x}_i^G) \\
\mathbf{x}_i^G & \text{otherwise}
\end{cases},
\]
where \(f(\cdot)\) is the objective function to be minimized.

\subsubsection{Potential advantages of Differential Evolution}
DE is potentially effective for music generation due to its real-valued vector representation and differential mutation. This potentially enables smooth transitions between generations, increasing the observable progress in training as a human repeatedly scores generated melodies. 

\subsubsection{Single user training}
Utilizing the concepts of melody generation and DE discussed here, one of the authors of this paper engaged in the iterative process of generating and ranking the melodies. Given the subjective nature of music, no quantitative metrics were employed to determine the notion of convergence. Instead, the stopping criterion was based on the user's qualitative assessment, concluding the process when the newly generated melodies were deemed to be of sufficient quality. 

After many generations of mutation, crossover, and selection, the population of musical vectors converged to a set of high-quality candidates, as deemed by the user. These vectors represent the "trained" state of the algorithm. The final population of vectors was then converted back into musical sequences, which are rendered as MIDI files or other musical formats. Then, the MIDI files were used to record instruments and synthesizers, serving as melodies in several songs. During this process, the following modulating features were incorporated to record the melodies. Some of the recordings were performed with advanced note-articulation MIDI keyboards, and others in the post-sound processing methods below. 

\begin{itemize}
    \item \textbf{Instrument Velocity Sensitivity}: The force applied to the keys was measured and modified to generate notes at varying velocities, allowing for dynamic sound expression.
    \item \textbf{Instrument Aftertouch Pressure}: Continuous pressure sensitivity after the initial strike of a note was utilized to modulate sound parameters, simulating effects such as breath pressure on wind instruments.
    \item \textbf{Instrument Lateral Motion Detection}: Motion of the keys was tracked to adjust pitch, enabling vibrato and pitch-bending effects.
    \item \textbf{Instrument Vertical Motion Detection}: Up and down motion along the keys was measured to modify sound properties such as brightness, texture, instrument inclusion, or depth.
    \item \textbf{Instrument Release Velocity}: The speed at which the keys were released was measured and modified to influence the resonance and decay of the sound.
\end{itemize}

These features are more likely to replicate the sounds of real instruments, because they capture the nuanced imperfections of finger actuation, such as variations in pressure and motion, which contribute to the authentic and expressive quality of live performances. In contrast, conventional computer-generated music often relies on discrete, on-off notes with full amplitude and minimal modulation, lacking the subtle dynamics that make real instruments sound lifelike. Although these effects were added after the melody DE generation, it would be possible to utilize DE to not only generate melodies, but also to generate key modulation.

\section{Results}

A total of six songs were created using these methods. Some of the songs predominantly feature melodies generated by DE, while others employ a hybrid approach, blending DE solutions with traditional, non-DE-assisted music generation. This was done intentionally to demonstrate the flexibility of incorporating DE methods into conventional production workflows. 

All six songs were sent to international record labels with the intent of receiving a release contract. Record labels, which receive a large quantity of songs, are incentivized to sign only tracks that are likely to be enjoyed by the community. Given the subjective nature of music and the difficulty in objectively assessing the quality of generated music, this approach aimed to demonstrate commercial viability by leveraging the expertise and market-driven motivations of industry professionals. The methods of production were not initially shared with the record labels to avoid potential bias resulting from the perceived strong or weak marketability of these unconventional methods. Notably, every submitted song received a signing contract opportunity. The generated songs will be released under the production name "Wondr Search" and can currently be listened to at \url{http://www.wondrsearch.com}.

Although the use of EAs for music generation is not necessarily novel, there appears to be a lack of industry case studies evaluating generated content by the music industry. Existing literature often focuses on novel generative methods rather than the commercial evaluation of said techniques. This study attempts to address this gap.

\section{Discussion} 
A remarkable and almost science fiction-like outcome of this single-user EA method is that the resulting vectors provide a measurable representation of the user's creative process and preferences. Essentially, the user's creativity and artistic inclinations are encoded within the DE vectors, allowing their stored creativity to be explored, manipulated, and expanded upon. These methods could be incorporated into existing Digital Audio Workstations, allowing generative melody plugins to learn from the user. Additionally, touring producers could use unique vector generations to create albums that are played exclusively in certain performances, resulting in truly unique, event-specific concert experiences. Unlike traditional ML, which relies on existing works and can therefore diminish the connection between the producer and the music, an EA approach preserves the touring producer's creative properties. This is likely to enhance the audience's future acceptance of computer-assisted content generation. 

This study underscores the potential of evolutionary generative algorithms in the realm of music production, providing a compelling glimpse into a future where math and human creativity can coexist. By integrating human feedback with EAs, we can capture the nuanced and dynamic qualities of human creativity, pushing the static boundaries of traditional AI-generated content. This approach preserves an artist's unique creative identity, therefore enhancing the diversity and cultural relevance of generated music. As the music industry continues to evolve, embracing these advanced generative techniques could lead to more innovative and personalized artistic expressions, offering listeners unique, performance-specific experiences. This exciting frontier supports a future where technology amplifies human creativity, fostering artistic exploration and discovery. 

Future research directions could include expanding the user base to further validate the generalizability of our findings. Additionally, exploring the application of DE algorithms in other creative domains, such as visual arts and literature, could provide valuable insights into the broader applicability of evolutionary generative methods.

 % \pagebreak

%Bibliography


\begin{thebibliography}{99}

\bibitem{davismoon2010combining}
S. Davismoon and J. Eccles, ``Combining musical constraints with Markov transition probabilities to improve the generation of creative musical structures,'' in \textit{European Conference on the Applications of Evolutionary Computation}, 2010, pp. 361–370.

\bibitem{doush2020automatic}
I. Abu Doush and A. Sawalha, ``Automatic music composition using genetic algorithm and artificial neural networks,'' in \textit{Malaysian Journal of Computer Science}, vol. 33, no. 1, 2020, pp. 35–51.

\bibitem{eck2002first}
D. Eck and J. Schmidhuber, ``A first look at music composition using LSTM recurrent neural networks,'' \textit{Istituto Dalle Molle Di Studi Sull’Intelligenza Artificiale}, vol. 103, p. 48, 2002.

\bibitem{farzaneh2019music}
M. Farzaneh and R. M. Toroghi, ``Music generation using an interactive evolutionary algorithm,'' in \textit{Mediterranean Conference on Pattern Recognition and Artificial Intelligence}, 2019, pp. 207–217.

\bibitem{horner1991genetic}
A. Horner and D. E. Goldberg, ``Genetic Algorithms and Computer-Assisted Music Composition,'' \textit{Proceedings of the International Computer Music Conference}, Urbana (Caracas, Venezuela), January 1991.


\bibitem{jacob1995composing}
B. L. Jacob, ``Composing with Genetic Algorithms,'' \textit{Proceedings of the International Computer Music Conference}, Banff, Alberta, September 1995.


\bibitem{makris2017combining}
D. Makris, M. Kaliakatsos-Papakostas, I. Karydis, and K. L. Kermanidis, ``Combining LSTM and feed forward neural networks for conditional rhythm composition,'' in \textit{International Conference on Engineering Applications of Neural Networks}, 2017, pp. 570–582.


\bibitem{mcvicar2014autoleadguitar}
M. McVicar, S. Fukayama, and M. Goto, ``AutoLeadGuitar: Automatic generation of guitar solo phrases in the tablature space,'' in \textit{2014 12th International Conference on Signal Processing (ICSP)}, 2014, pp. 599–604.

\bibitem{abbou2020deepclassic}
R. Abbou, ``DeepClassic: Music Generation with Neural Networks,'' Stanford CS224N Custom Project, in \textit{Computational and Mathematical Engineering, Stanford University}, 2020.

\bibitem{mishra2019long}
A. Mishra, K. Tripathi, L. Gupta, and K. P. Singh, ``Long short-term memory recurrent neural network architectures for melody generation,'' in \textit{Soft Computing for Problem Solving}, Springer, 2019, pp. 41–55.

\bibitem{shao2019how}
Y. Shao, C. Zhang, J. Zhou, T. Gu, and Y. Yuan, ``How does culture shape creativity? A mini-review,'' \textit{Frontiers in Psychology}, vol. 10, pp. 1219, May 2019.

\bibitem{farzaneh2024gga}
M. Farzaneh and R. M. Toroghi, ``GGA-MG: Generative Genetic Algorithm for Music Generation,'' Media Engineering Faculty, Iran Broadcasting University, Tehran, Iran, 2024.

\bibitem{tokui2000music}
N. Tokui and H. Iba, ``Music composition with interactive evolutionary computation,'' in \textit{Proceedings of the Third International Conference on Generative Art}, vol. 17, 2000, pp. 215–226.

\bibitem{otoole2023novelty}
K. O’Toole and E. Horvát, ``Novelty and cultural evolution in modern popular music,'' \textit{EPJ Data Science}, vol. 12, no. 3, 2023.

\bibitem{Smith2018}
S. Smith and B. Samples, ``Music through a mathematical lens,'' \textit{gcsu.edu}, 2018.

\bibitem{Anderling2014}
V. Anderling, O. Andreasson, C. Olsson, S. Pavlov, C. Svensson, and J. Wikner, ``Generation of music through genetic algorithms,'' \textit{Chalmers University of Technology, University of Gothenburg}, June 2014.

\bibitem{StornPrice1997}
R. Storn and K. Price, ``Differential evolution – A simple and efficient heuristic for global optimization over continuous spaces,'' \textit{Journal of Global Optimization}, vol. 11, pp. 341–359, 1997.


\end{thebibliography}
\end{document}